\documentclass[10pt,twocolumn,a4paper]{article}

\usepackage[final]{cvww} 

\usepackage[pagebackref,breaklinks,colorlinks,allcolors=cvwwblue]{hyperref}
\usepackage[linesnumbered,ruled,vlined]{algorithm2e}
\usepackage{makecell}
\usepackage{multirow}

\usepackage{algorithm}
\usepackage{amsmath} 
\usepackage{amsfonts} 
\usepackage{amssymb} 
\usepackage{algorithmic}
\usepackage{tabularx}
\newcolumntype{Y}{>{\centering\arraybackslash}X}
\usepackage{arydshln}
\usepackage{graphicx}
\def\paperID{20} 

\title{A Data-Centric Approach to 3D Semantic Segmentation of Railway Scenes}

\author{Nicolas Münger \quad Max Ronecker \quad Xavier Diaz \quad Michael Karner \\[0.3em]
SETLabs Research GmbH\\
Elsenheimerstraße 55, 80687 München, Germany\\
{\tt\small \{firstname.lastname\}@setlabs.de}
\and
Daniel Watzenig\\
Graz University of Technology\\
Inffeldgasse 16/II, 8010 Graz, Austria\\
{\tt\small daniel.watzenig@tugraz.at}
\and
Jan Skaloud\\
EPFL - Swiss Federal Technology Institute of Lausanne\\
GR A2 392 (Bâtiment GR) , 1015 Lausanne , Switzerland\\
{\tt\small jan.skaloud@epfl.ch}
}

\begin{document}
\maketitle


\begin{abstract}
LiDAR-based semantic segmentation is critical for autonomous trains, requiring accurate predictions across varying distances. This paper introduces two targeted data augmentation methods designed to improve segmentation performance on the railway-specific OSDaR23 dataset. The person instance pasting method enhances segmentation of pedestrians at distant ranges by injecting realistic variations into the dataset. The track sparsification method redistributes point density in LiDAR scans, improving track segmentation at far distances with minimal impact on close-range accuracy. Both methods are evaluated using a state-of-the-art 3D semantic segmentation network, demonstrating significant improvements in distant-range performance while maintaining robustness in close-range predictions. We establish the first 3D semantic segmentation benchmark for OSDaR23, demonstrating the potential of data-centric approaches to address railway-specific challenges in autonomous train perception.
\end{abstract}

\section{Introduction}
\label{sec:intro}

Rail transport offers a sustainable alternative to other transportation modes, emitting significantly lower carbon emissions~\cite{deutschebahnMetropolitanNetworkStrong2023}. Its continued development, especially through autonomous train operation (ATO), is critical to achieving climate goals like those in the European Union's Green Deal. ATO, defined from GoA0 (manual) to GoA4 (fully automated)~\cite{rizziAutomatedMetros}, addresses labor shortages, increases operational flexibility and reliability, and optimizes service frequency. The Lausanne metro M2 line, a GoA4 system, demonstrates these benefits through higher frequency and adaptability. However, while fully automated systems work well in controlled settings, such as metro lines, implementing GoA3–4 in open rail networks is challenging due to unpredictable obstacles and the absence of physical barriers. Ensuring safety in open rail ATO is therefore a key research area.

Robust perception systems are essential for obstacle detection and hazard identification in ATO. LiDAR (Light Detection and Ranging) suits these tasks by providing rich 3D geometric information~\cite{qiPointNetDeepLearning2017}. LiDAR semantic segmentation, assigning a class to each 3D point, enables detailed environmental understanding. In autonomous driving, 3D object detection~\cite{Chen_2016_CVPR, simonelli_ICCV_2021, charles_CVPR_2017, Shi_2019_CVPR, charles_frustum_CVPR_2018, liu2022bevfusion} and semantic segmentation~\cite{Badrinarayanan_2017_PAMI,mohan2021ijcv,yanwei_2019,Sirohi_2022,zhangReviewDeepLearningBased2019} are well-studied across many modalities. However, applying these techniques to autonomous train operation has received less attention, partly due to limited public datasets. The OSDaR23 dataset~\cite{tagiewOSDaR23OpenSensor2023} addresses this gap by providing data for various railway perception tasks. This paper applies deep learning-based 3D semantic segmentation to LiDAR point clouds in the railway domain using OSDaR23. We focus on safety-critical classes, emphasizing long-range segmentation accuracy due to trains’ substantial braking distances. We also adopt a data-centric approach, introducing domain-specific data augmentations to improve robustness and performance.

\begin{figure}[t]
\centering
\includegraphics[width=.95\linewidth]{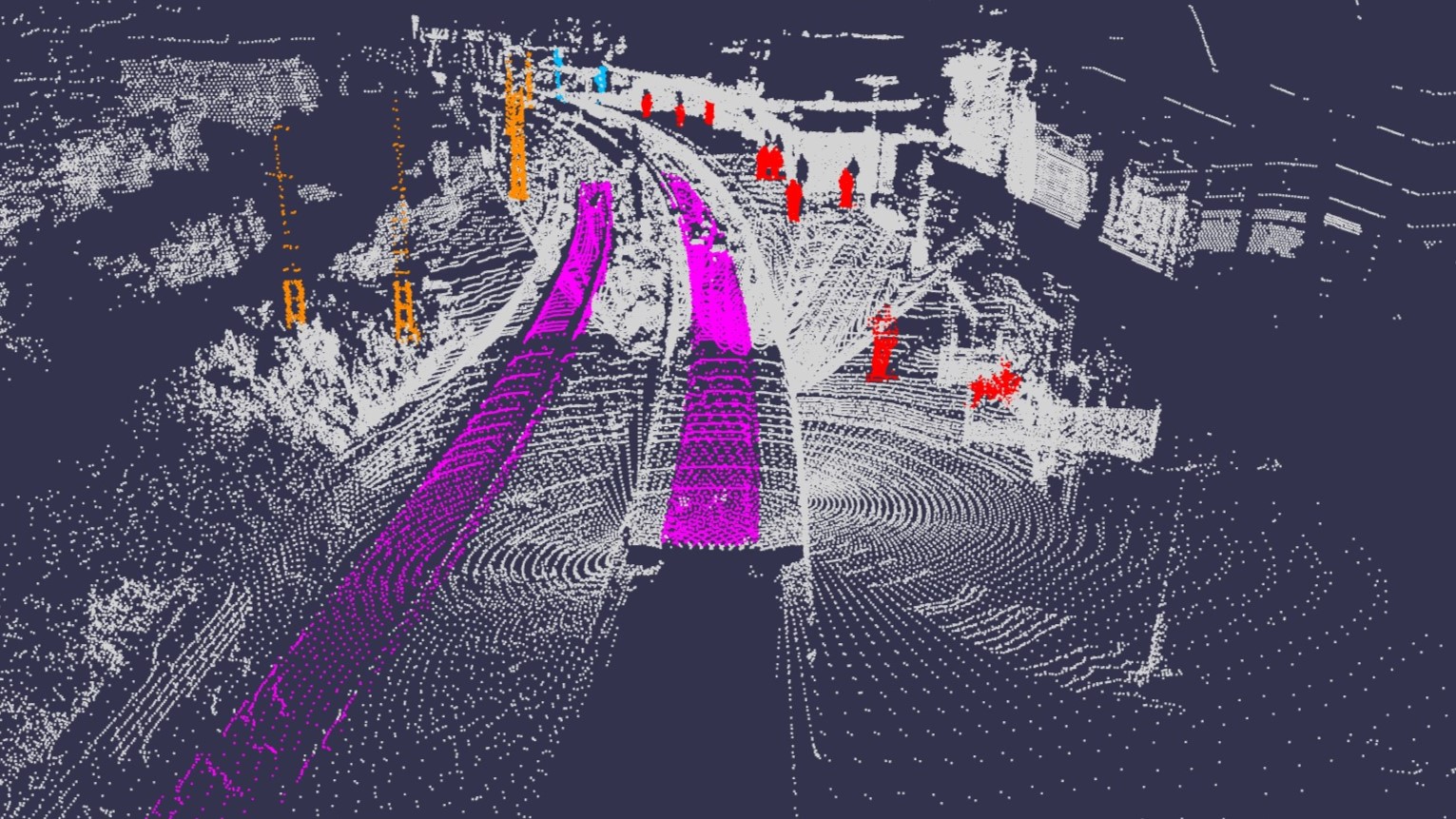}
\caption{Example of a segmented pointcloud from the OSDaR23 dataset~\cite{tagiewOSDaR23OpenSensor2023}}
\label{fig:example_osdar_23}
\end{figure}

\textbf{Contributions} \\
This paper introduces targeted data augmentation methods for LiDAR semantic segmentation in the railway domain, evaluated on the real-world OSDaR23 dataset.
\begin{enumerate}
    \item Comprehensive evaluation of a state-of-the-art 3D semantic segmentation network on OSDaR23, including dataset analysis.
    \item A person instance pasting augmentation method to enhance pedestrian segmentation at distant ranges.
    \item A track sparsification augmentation method to improve track segmentation by redistributing point density.
    \item Report the first 3D semantic segmentation results on the OSDaR23 dataset.
\end{enumerate}

\section{Background}
\label{sec:background}
This background section provides a general overview of point cloud segmentation, followed by segmentation and augmentation techniques specific to the railway domain.

\subsection{Point cloud semantic segmentation}\label{section:pc_sem_seg}
Semantic segmentation assigns a class label to each element of the input. While image-based segmentation assigns labels to pixels, point cloud segmentation must handle unordered, unstructured 3D points. Deep learning has become the standard approach, surpassing traditional techniques~\cite{xieLinkingPointsLabels2020}. Methods are typically categorized into view-based, voxel-based, and point-based approaches, each imposing structure onto the raw data differently.

\subsubsection*{View-based methods} 
View-based methods project the point cloud into one or multiple 2D images, leveraging established image-based segmentation. SnapNet~\cite{boulchUnstructuredPointCloud2017} generates RGB-depth snapshots from various viewpoints, applies a CNN for labeling, and back-projects labels to 3D. CENet~\cite{chengCenetConciseEfficient2022} uses spherical projection and channels $(x,y,z,d,r)$ for each pixel. Larger image widths improve performance but slow inference. However, these methods lose some 3D geometric fidelity due to projection.

\subsubsection*{Voxel-based methods} 
Voxel-based methods discretize the point cloud into a volumetric grid and apply 3D CNNs. PVKD~\cite{houPointtoVoxelKnowledgeDistillation2022}, for example, builds on Cylinder3D~\cite{zhuCylindricalAsymmetrical3D2021a} and employs a teacher-student framework, achieving similar accuracy at lower latency. Despite structuring the data, voxelization introduces resolution limits and can demand high memory.

\subsubsection*{Point-based methods} 
Point-based methods directly process points without explicit restructuring. PointNet~\cite{qiPointNetDeepLearning2017} introduced MLP-based features and max-pooling for permutation invariance. Transformers, as in Point Transformer~\cite{zhaoPointTransformer2021} and its improved PTV3~\cite{wuPointTransformerV32023}, leverage self-attention for robust performance. This preserves data fidelity but can be slower.

In summary, view-based and voxel-based methods effectively impose structure at the cost of fidelity, while point-based methods maintain full data integrity but may be computationally more demanding.

\begin{figure}[t]
    \centering
    \textbf{View-based}\\[3pt]
    \includegraphics[width=\linewidth, trim={0 1.5cm 0 19cm},clip]{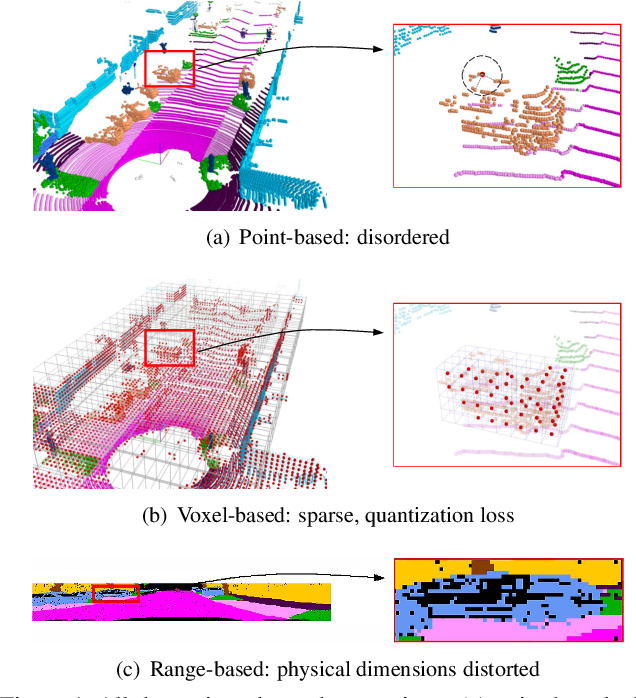}\\[10pt]

    \textbf{Voxel-based}\\[3pt]
    \includegraphics[width=\linewidth, trim={0 8cm 0 10cm},clip]{Figures/Literature_review/RPVNET_explanation_figure.png}\\[10pt]

    \textbf{Point-based}\\[3pt]
    \includegraphics[width=\linewidth, trim={0 17cm 0 0},clip]{Figures/Literature_review/RPVNET_explanation_figure.png}

    \caption{Schematic representation of three main deep learning-based methods for semantic segmentation of point cloud data. Adapted from~\cite{xuRPVNetDeepEfficient2021}.}
    \label{fig:vis_dl_PCSS_methods}
\end{figure}

Fig. \ref{fig:vis_dl_PCSS_methods} shows an example for each of the three approaches.

\subsection{Railway-domain focused segmentation}
Prior work on railway point cloud segmentation focused mainly on infrastructure inspection. ~\cite{soilanSEMANTICSEGMENTATIONPOINT2020} segmented tunnel scenes into ground, lining, wiring, and rails using KPConv~\cite{thomasKPConvFlexibleDeformable2019} and PointNet~\cite{qiPointNetDeepLearning2017}. Similarly,~\cite{grandioPointCloudSemantic2022} employed a PointNet++~\cite{qiPointNetDeepHierarchical2017}-based architecture to classify rails, cables, and traffic signals. These efforts used non-public datasets and older architectures, and did not target autonomous train operation.

In contrast, the automotive field has benefited from large-scale, publicly available datasets like Waymo Open Dataset~\cite{sunScalabilityPerceptionAutonomous2020}, nuScenes~\cite{caesarNuScenesMultimodalDataset2020}, and SemanticKITTI~\cite{behleySemanticKITTIDatasetSemantic2019}. Comparable resources remain scarce in the railway domain. Existing sets, such as WHU-Railway3D ~\cite{dongRegistrationLargescaleTerrestrial2020} and Rail3D~\cite{kharroubiMultiContextPointCloud2024}, focus on infrastructure and rely on multi-frame reconstructions, not reflecting real-time conditions. OSDaR23~\cite{tagiewOSDaR23OpenSensor2023} addresses this gap with single-frame LiDAR data and classes relevant to autonomous rail operation, enabling models tailored to open railway environments.

\subsection{Data augmentation methods for point clouds}
Data-centric AI aims to enhance model performance by improving data quality and diversity rather than solely refining architectures. In point cloud segmentation, data augmentation (DA) introduces variations—such as rotations, translations, and sparsifications—to enrich training data and improve generalization \cite{zhuAdvancementsPointCloud2024,chengImproving3DObject2020,ngiamStarNetTargetedComputation2019}.

Part-aware augmentation \cite{choiPartAwareDataAugmentation2021} applies transformations to specific object regions (e.g., sparsifying parts of cars or pedestrians), reducing reliance on dense shapes and aiding recognition at longer distances. PolarMix \cite{xiaoPolarMixGeneralData2022} integrates entire LiDAR scans by angular swapping or instance-level rotate-pasting, increasing variability at both scene and object levels. Both methods have demonstrated notable performance gains in 3D tasks and inspire the DA techniques explored in this work.

\section{Initial Analysis}

In this section, we evaluate the baseline performance of Point Transformer V3 (PTV3) on the OSDaR23 dataset. Since the dataset has seen limited use in prior research, its suitability for semantic segmentation tasks, along with potential performance bottlenecks, remains unclear. This analysis aims to establish a baseline understanding of the model's strengths and limitations, highlighting key challenges such as class imbalance and long-range prediction issues. These findings will guide subsequent efforts to enhance model performance through targeted improvements.

\subsection{Baseline}\label{section:baseline}
For our baseline, we require a modern, high-performing semantic segmentation model suited for LiDAR point clouds. Point Transformer V3 (PTV3)\cite{wuPointTransformerV32023} is the current top performer on the SemanticKITTI benchmark, demonstrating strong segmentation accuracy with reasonable inference speed. Although relatively new and less cited, it builds on the widely adopted Point Transformer\cite{zhaoPointTransformer2021,wuPointTransformerV22022} architecture, making it a robust choice for our experiments.

\subsection{Experiment setup}\label{section:experiment_setup}
We conduct our experiments on OSDaR23 \cite{tagiewOSDaR23OpenSensor2023}, a single-frame, multi-sensor LiDAR dataset tailored for open-rail scenarios. While the dataset includes many annotated classes, several have very few points, leading to class imbalance. To address this, we merge and discard some classes as shown in Table \ref{tab:class_mapping}. Overlapping annotations, such as \textit{switch} on \textit{track}, are removed to avoid confusion.

\begin{table}[t]
\centering
\caption{Class mapping for OSDaR23.}
\begin{tabular}{l|l}
\hline
\textbf{Original classes} & \textbf{Mapped class} \\ \hline
person, crowd & person \\
train, wagons & train \\
bicycle, animal, signal\_bridge & background \\
transition, track & track \\
road\_vehicle & road\_vehicle \\
catenary\_pole & catenary\_pole \\
signal\_pole, signal & signal \\
buffer\_stop & buffer\_stop \\
switch & discarded \\ \hline
\end{tabular}
\label{tab:class_mapping}
\end{table}

We adapt data augmentation to the forward-facing LiDAR viewpoint, limiting global rotations and flips. Sensor-specific intensity normalization compensates for varying hardware sources. We train Point Transformer V3 (PTV3) using cross-entropy and Lovász-Softmax loss \cite{bermanLovaszSoftmaxLossTractable2018}, and set the learning rate to 0.001. All experiments use the official train, validation, and test splits to maintain consistency.

\subsection{Baseline Performance}
We begin by examining the baseline model’s overall segmentation performance on the validation set. As shown in Table \ref{Table:IoU_baseline_exp}, the model (PTV3) achieves a mean IoU (mIoU) of 74.49\%, indicating solid overall accuracy across classes. However, this summary metric masks performance issues at longer ranges.

\begin{table*}[t]
\centering
\caption{Summary results for the baseline experiment. (Validation dataset)}
\begin{tabular}{cccccccc|c}
\hline
\multicolumn{8}{c|}{\textbf{IoU}} & \textbf{mIoU} \\ 
\hline
\textbf{background} & \textbf{person} & \textbf{train} & \textbf{road vehicle} & \textbf{track} & \textbf{catenary pole} & \textbf{signal} & \textbf{buffer stop} & \textbf{Overall} \\ 
\hline
96.84 & 69.65 & 86.39 & 70.09 & 82.89 & 47.40 & 48.80 & 93.86 & 74.49 \\
\hline
\end{tabular}
\label{Table:IoU_baseline_exp}
\end{table*}

\begin{figure}[t]
\centering
\includegraphics[width=\linewidth]{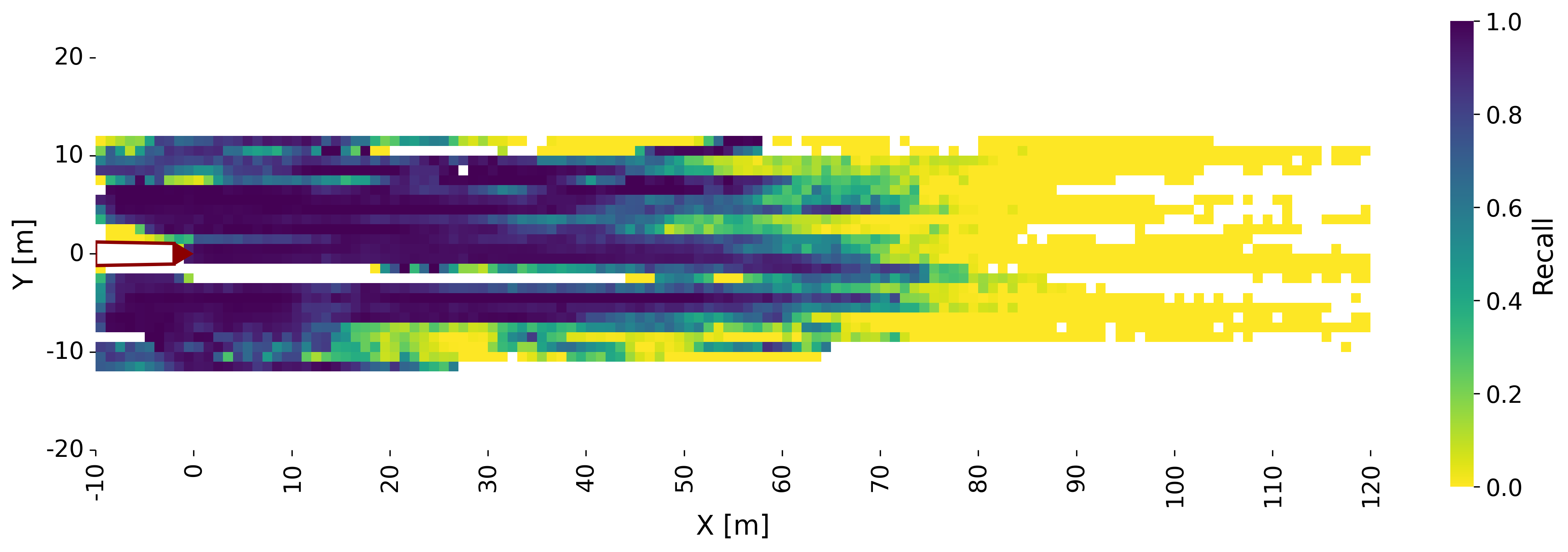}
\caption{Recall for the class track across the validation set. High recall is observed close to the sensor, with performance decreasing beyond 60 m.}
\label{fig:track_recall}
\end{figure}

Fig. \ref{fig:track_recall} shows the recall map for the class track. For each planar grid cell of 1x1 meter, the recall is computed. The values are obtained over all frames of the validation set, providing an overview of the performances given the spatial location. In the ranges close to the sensor the recall is generally high. Beyond x=60m, however, the recall quickly degrades. This means the network has good capabilities at identifying the track points at close range but misses points further. 

Similarly, person segmentation suffers at longer ranges, as reflected in the range IoU (rIoU) results (Table \ref{table:person_baseline_analysis}). Although performance is strong at mid-range (40--60 m), it drops significantly beyond 60 m. This decline correlates with fewer training samples at longer distances, indicating that data scarcity limits long-range accuracy.
\begin{table}[t]
\centering
\caption{Baseline range-based IoU for the person class and approximate number of training instances.}
\begin{tabular}{c|c|c}
\hline
Distance range & IoU [\%] (Val) & \#Instances (Train) \\ \hline
0--20 m & 80.40 & \(\approx 5900\) \\
20--40 m & 69.73 & \(\approx 3500\) \\
40--60 m & 81.23 & \(\approx 500\) \\
60--80 m & 31.36 & \(\approx 350\) \\
80--100 m & 45.17 & \(\approx 100\) \\ \hline
\end{tabular}
\label{table:person_baseline_analysis}
\end{table}

In summary, while the baseline model performs well overall, it struggles to maintain performance at longer distances for key classes like track and person. Insufficient training data in these ranges is a likely contributor to weaker performance, motivating the need for data augmentation and other strategies to improve long-range segmentation results.

%
%
%
%
\section{Methodology}

This section outlines the data-centric strategies developed to address the dataset-related limitations identified in the baseline analysis. Our methodology focuses on two key augmentations: track sparsification and person instance pasting, tailored to the characteristics of the OSDaR23 dataset.

\subsection{Tracks sparsification}
Building on the part-aware data augmentation method \cite{choiPartAwareDataAugmentation2021}, a new strategy was developed to improve track prediction accuracy at farther ranges.

Dense parts of track instances are sparsified by adapting the number of points per range for each track instance. The goal is to equalize point density by reducing points near the sensors to match the density farther away. This is achieved by evaluating the number of points within a window of width \( W \) at a distance \( d \) from the origin, where \( C_{max} \) represents the point count in the farthest range. Closer ranges are then randomly downsampled to match \( C_{max} \), ensuring uniform density.

Let \( P_{\text{track}, i[d-W,d]} \)  denote the set of points belonging to the \( i^{\text{th}} \) track instance in the planar distance range \([d-W, d]\),. The variables \( W \) (window width) and \( C_{max} \) can be adjusted based on sensor specifications and use case requirements. The pseudocode for the transformation is provided in Algorithm \ref{alg:track_sparsification}.

\begin{algorithm}[b]
\caption{Track Instance Sparsification}\label{alg:track_sparsification}
\begin{algorithmic}
\STATE \textbf{Input:} $P_{t,i}$ (points of track $i$), $d_{max}$ (upper range), $W$ (window width)
\STATE \textbf{Output:} Downsampled $P_{t,i}$

\STATE $D_i \gets$ planar distances from origin for $P_{t,i}$
\STATE $d_{max} \gets \min(d_{max}, \max(D_i))$
\STATE $C_{max} \gets$ count points in $[d_{max}-W, d_{max})$

\WHILE{$d_{max} > 0$}
    \STATE $d_{max} \gets d_{max} - W$
    \STATE $C \gets$ count points in $[d_{max}-W, d_{max})$
    \IF{$C > C_{max}$}
        \STATE Remove $C - C_{max}$ points from $P_{t,i}$
    \ENDIF
\ENDWHILE

\STATE \textbf{Return} $P_{t,i}$
\end{algorithmic}
\end{algorithm}

\begin{figure}[b]
\centering
\includegraphics[width=1.\linewidth]{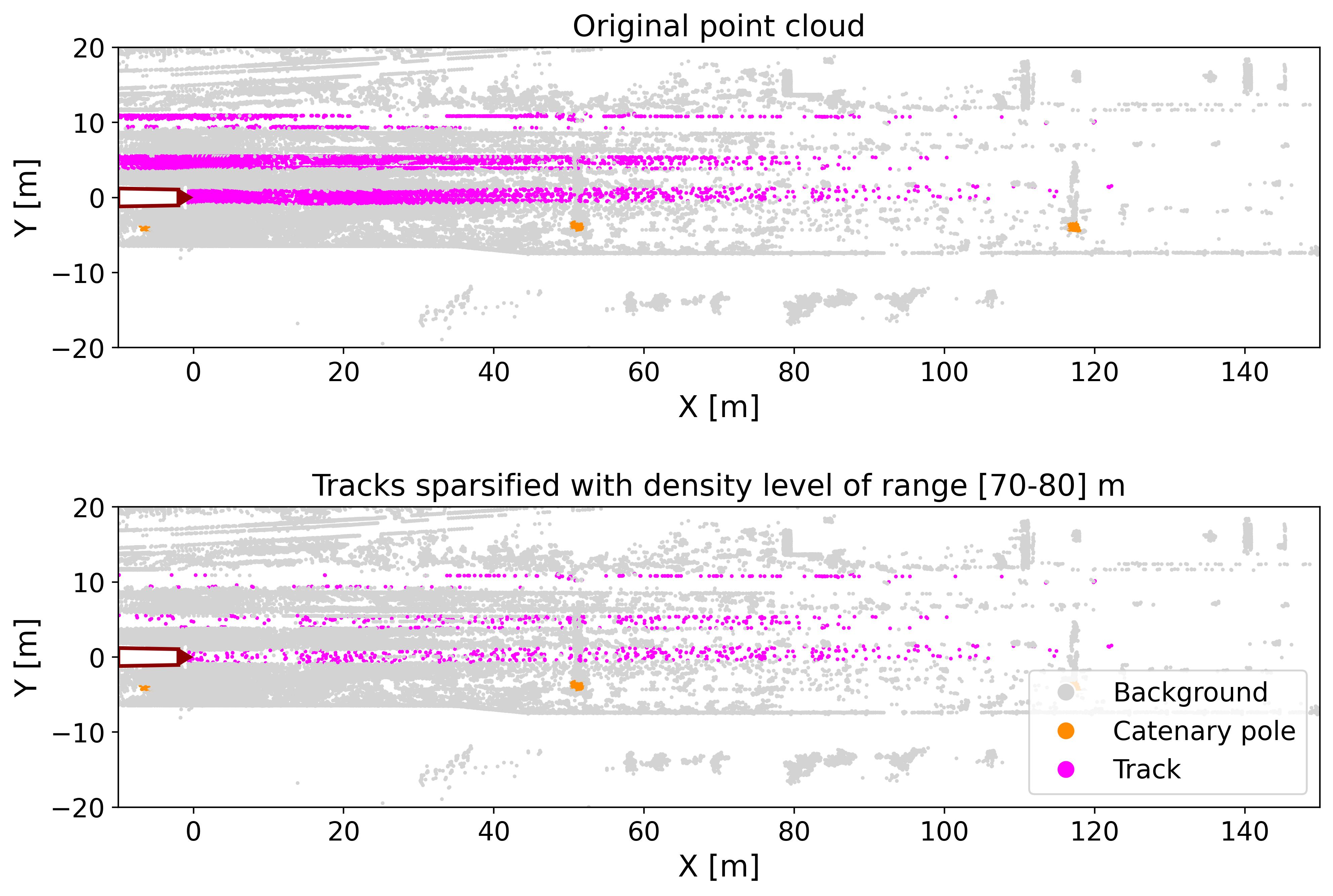}
\caption{Effect of the tracks sparsification transformation on scene 3\textunderscore fire\textunderscore site\textunderscore 3.1, frame 58 from the OSDaR23 dataset.}
\label{fig:sparse_aug_vis}
\end{figure}

This procedure is applied to all track instances in a frame. Fig. \ref{fig:sparse_aug_vis} shows a point cloud before and after the transformation. In this example, the window width \( W \) is set to 10 meters, and \( C_{\text{max}} \) is set to 80 meters. The desired density is determined within the range \([C_{\text{max}} - W, C_{\text{max}}]\) (70–80 meters). Points beyond 70 meters remain unchanged, while those closer than 70 meters are significantly downsampled.

\subsection{Person Instance Pasting}

Inspired by PolarMix \cite{xiaoPolarMixGeneralData2022}, we developed a methodology to paste person instances from one frame into another during training. This approach diversifies pedestrian samples by increasing their population. Unlike PolarMix, where objects are rotated around the vehicle without individual transformations, our method accounts for the forward-facing point clouds in OSDaR23, which differ from the 360-degree coverage in datasets like SemanticKITTI. A simple rotation would place instances outside the field of view, necessitating significant adaptation of the original methodology.

As in PolarMix, Scan \textit{A} denotes the frame undergoing transformation, and Scan \textit{B} denotes the randomly selected frame from the training set containing at least one person instance.

Each instance of Scan \textit{B} goes through a set of individual transformations, applied in this order:
\begin{enumerate}[topsep=0pt,itemsep=-1ex,partopsep=1ex,parsep=1ex]
\item Flipping along the X axis with 0.5 probability.
\item Random rotation around the instance's center along the Z axis, within the range [-180°, 180°]. 
\item Random shift along the Y axis, within the range [-2m, 2m]. 
\item Random shift towards the back of the scene, along the X axis. \label{listitem:X_shift}
\item Shifting along the Z axis so as to be at a realistic height. \label{listitem:Z_shift}
\end{enumerate}

\begin{figure}[t]
\centering
\includegraphics[width=\linewidth]{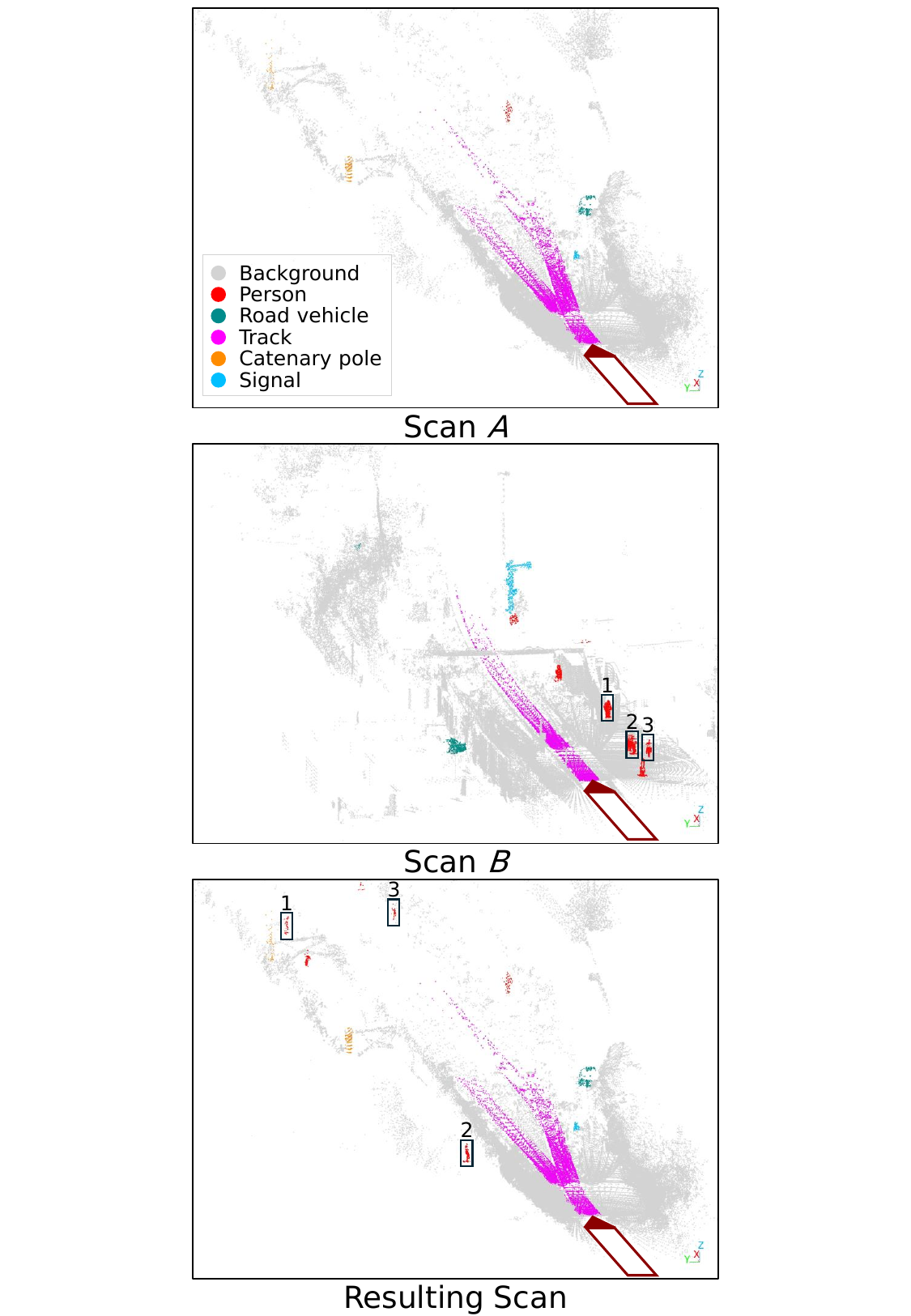}
\caption{Visualisation of the person instances pasting transformation. Best viewed zoomed in.}
\label{fig:Pedestrian_paste_example}
\end{figure}

An example for scan A and B and the produced result is shown in Fig.~\ref{fig:Pedestrian_paste_example}.

For the X-axis shift, instances are translated further from the sensor to balance the distribution, with density adjustments based on the histogram of points per instance. The instance is downsampled to match the expected point count \( N \), sampled randomly within \([N - 0.1N, N + 0.1N]\).

For the Z-axis shift, instances are adjusted to align with the ground. The ground height is estimated as the mean height of points in Scan \textit{A} under the instance's bounding box. Special cases include estimating the ground height from railway tracks when no points overlap or ignoring unrealistic heights (e.g., above 150 cm).

\begin{figure}[bt]
\centering
\includegraphics[width=\linewidth]{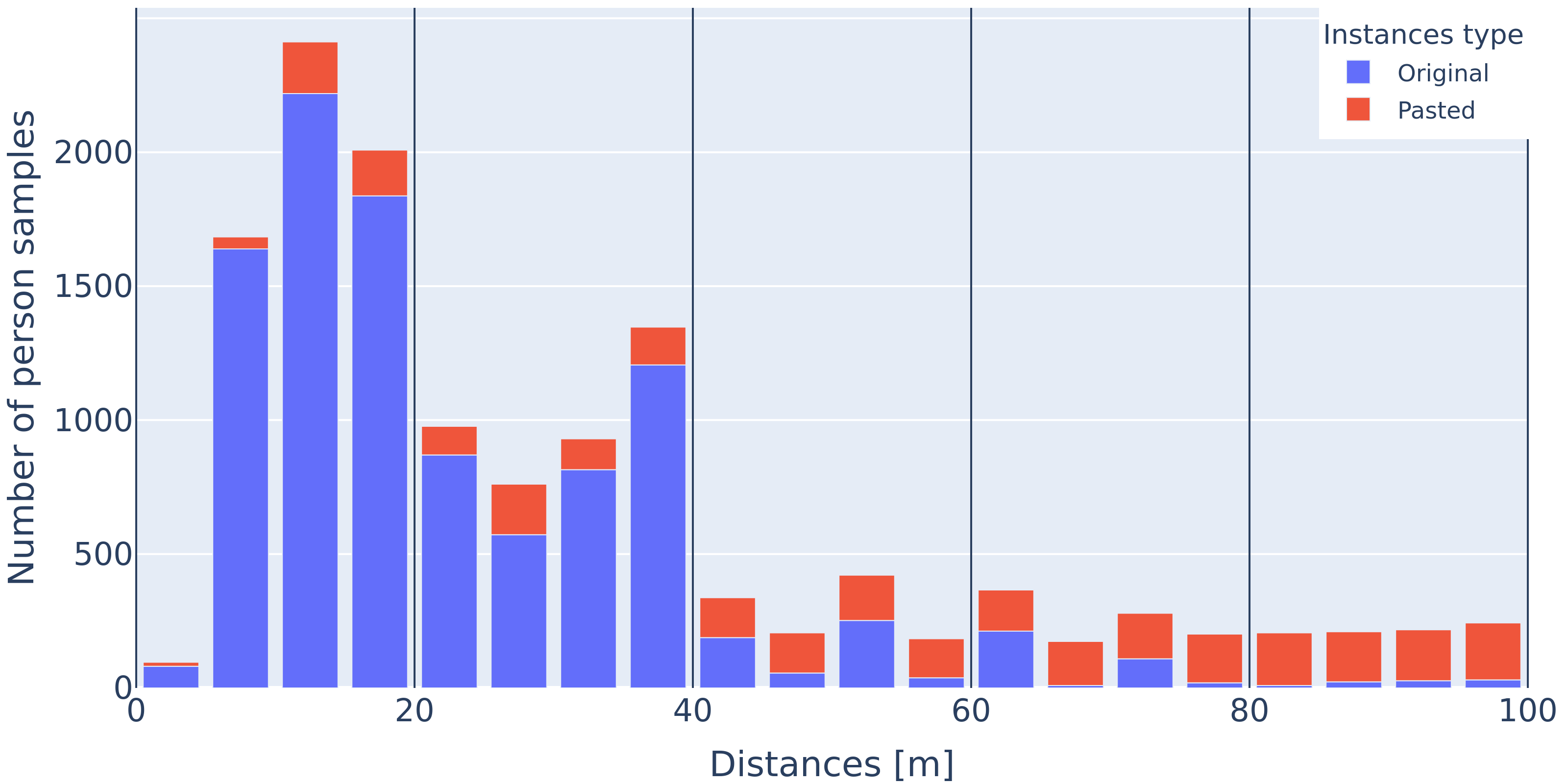}
\caption{New distribution of samples with the person instance pasting DA applied on all frames from the train set.}
\label{fig:pasted_new_distribution}
\end{figure}

After applying the transformation, the augmented dataset shows a more balanced distribution of person instances across ranges, particularly in previously sparse areas as shown in Fig~\ref{fig:pasted_new_distribution}.

\section{Results}
This section presents the results of applying the data augmentation (DA) methods during training, with varying proportions of affected samples. Models are first evaluated on the validation set to select the best for each task, which are then tested on the test set.

To reduce the foreground bias of IoU, we propose the mean range IoU (mean rIoU), which assigns equal importance to IoUs across all ranges. Let $\mathrm{rIoU}_i$ represent the range IoU for bin $i$. The mean rIoU is defined as:

\begin{equation}
  \mathrm{mean \ \ rIoU} = \frac{1}{N} \sum_{i=1}^{N} \mathrm{rIoU}_i
  \label{eq:mean_rIoU}
\end{equation}

where $\mathrm{rIoU}_i$ is computed for points in the range $[r_{min,i}, r_{max,i}[$, with $r_{min,i}$ and $r_{max,i}$ as bin boundaries, and $N$ as the number of bins.

\subsection{Track sparsification}

This section evaluates the impact of the track sparsification DA method, tested with two density selection distances (DSD): 70-80m and 40-50m. The augmentation was applied with varying probabilities (\( p \)) during training, with range IoUs computed at 20m intervals from 0-100m. The baseline corresponds to \( p=0 \) (no augmentation), while \( p=1 \) applies the transformation to all training samples.

The ablation study identifies the best augmentation probabilities as \( p=0.6 \) for DSD 70-80m and \( p=0.9 \) for DSD 40-50m. Table \ref{table:summary_val} summarizes the results. The model with DSD 40-50m at \( p=0.9 \) achieves the highest mean rIoU (59.49\%), improving performance in ranges 40-60m and 60-80m by over 7 percentage points compared to the baseline. Both augmented models show improvements in the farthest range (80-100m), while maintaining strong performance near the origin. The baseline achieves the highest rIoU in 0-20m but with minimal difference (0.01 percentage points).

\begin{figure}[b]
\centering
\includegraphics[width=1\linewidth]{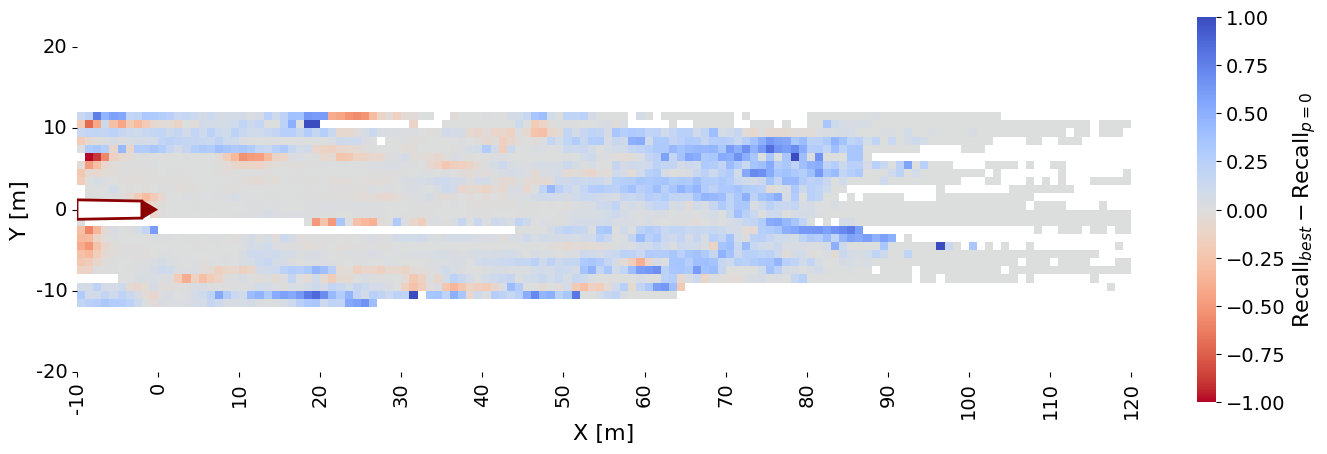}
\caption{Recall difference between the best model and model with no augmentation on the validation set.}
\label{fig:diff_recall_map_val_22_05}
\end{figure}

\begin{table*}[t]
\caption{Summary metrics for baseline and best models of track sparsification and person instance pasting (validation set)}
\centering
\begin{tabular}{lcccccc}
\hline
                                   & mean rIoU & r0-20 & r20-40 & r40-60 & r60-80  & r80-100 \\  [0.5ex] \hline \hline
\multicolumn{7}{c}{\textbf{Track Sparsification (Density Selection Distances)}} \\ \hline
\multicolumn{1}{l|}{baseline}      & 56.52 & \textbf{86.76} & 82.05 & 64.98 & 40.98   & 7.82    \\
\multicolumn{1}{l|}{70-80m (best)} & 58.01 & 86.64 & 81.61 & 64.74 & 43.15   & \textbf{13.93}   \\
\multicolumn{1}{l|}{40-50m (best)} & \textbf{59.49} & 86.75 & \textbf{82.19} & \textbf{66.70} & \textbf{48.29}   & 13.50   \\ \hline
\multicolumn{7}{c}{\textbf{Person Instances Pasting}} \\ \hline
\multicolumn{1}{l|}{baseline}      & 61.57 & 80.40  & 69.73 & \textbf{81.23} & 31.36   & 45.17   \\
\multicolumn{1}{l|}{online (best)} & \textbf{66.99} & 78.66  & \textbf{70.12} & 78.49 & \textbf{49.92}   & 57.76\\
\multicolumn{1}{l|}{offline (best)} & 66.77 & \textbf{80.98}  & \textbf{70.12}  & 79.46 & 44.59   & \textbf{58.70 } \\ \hline
\end{tabular}
\label{table:summary_val}
\end{table*}
The selected model (DSD 40-50m, \( p=0.9 \)) also improves recall at farther distances, as shown in Fig. \ref{fig:diff_recall_map_val_22_05}, while maintaining comparable performance closer to the origin. The results demonstrate that the track sparsification DA method effectively enhances performance at greater distances when applied with the identified optimal probabilities.

\subsection{Person instances pasting}

This section evaluates the impact of the person instances pasting DA method using two approaches: online augmentation and offline dataset inflation.
Online augmentation applies transformations to training samples in real-time, modifying data on-the-fly during training. Offline augmentation pre-processes the dataset by adding transformed samples, increasing its size before training. For person instance pasting, online augmentation randomly pastes instances during training, while offline augmentation generates augmented frames beforehand and incorporates them into the dataset.

In online augmentation, the probability (\(p\)) determines the likelihood of applying transformations to a sample during each training iteration. Higher \(p\) dynamically increases the number of augmented samples in each epoch.

In offline augmentation, the dataset size is expanded by adding transformed samples, controlled by the augmentation ratio (\(\alpha\)). For instance, \(\alpha=1.0\) doubles the dataset by adding a transformed version of each sample, while \(\alpha=0.5\) increases the size by 50\%.

Again an ablation study is conducted to determine the optimal values for \(p\) and \(\alpha\). The best models are selected based on mean rIoU: \(p=0.8\) for online DA and \(\alpha=0.1\) for offline DA. Table \ref{table:summary_val} compares these models with the baseline. Both approaches show significant improvements in the farthest ranges (60-100m). The online method achieves an 18.56 percentage-point increase in range 60-80m and a 12.59-percentage-point increase in range 80-100m over the baseline. Similarly, the offline method improves range 80-100m by 13.53 percentage-point. For closer ranges (0-60m), the differences are minimal, with variations below 3 percentage points. The online DA trained model (\(p=0.8\)) achieves the highest mean rIoU (66.99\%) and is the overall best model for this task.

\subsection{Results on Test Set}

The best-performing models identified during validation were evaluated on the test set to assess their generalization to new data.

\begin{table*}[t]
\centering
\caption{Summary of test set results. TS: track sparsification, PIP: person instance pasting (online). For each method, the best-performing model from the validation set is used.}
\begin{tabular}{lcccccccc|c}
\hline
& \multicolumn{8}{c|}{\textbf{IoU}}             & \textbf{mIoU} \\ [0.1cm]
& background & person & train & \makecell{road \\ vehicle} & track & \makecell{catenary \\ pole} & signal & \makecell{buffer \\ stop} & \\ [0.5ex] \hline \hline
\multicolumn{1}{c|}{Baseline} & \textbf{97.09}      & \textbf{77.98}  & \textbf{59.87} & 72.06        & 81.29 & 71.01         & \textbf{56.83}  & 0.53       & 64.58 \\

\multicolumn{1}{c|}{TS (best)} & 97.03      & 77.27  & 57.33 & 73.51        & 80.60 & 75.67         & 53.27  & 0.29       & 64.37 \\

\multicolumn{1}{c|}{PIP online (best)} & 97.02      & 77.21  & 57.47 & \textbf{77.33}        & \textbf{81.34} & \textbf{75.83}        & 52.25  & \textbf{0.81}       & \textbf{64.91} \\

\hline
\end{tabular}
\label{Table:IoU_comparison_test_set}
\end{table*}

\subsubsection{Class Track}
The best model for track sparsification (TS, DSD 40-50m, \(p=0.9\)) improves rIoUs in ranges beyond 40m, with a 5 percentage-point increase in 80-100m compared to the baseline. However, a slight decrease in the 0-20m range is observed, attributed to the network focusing on sparsified far-range points during training, potentially neglecting the dense close-range regions. Recall maps show significant gains in 60-90m, reflecting better far-range detection, while closer ranges see some localized recall reduction on the locomotive's sides.

\subsubsection{Class Person}
The best model for person instance pasting (PIP online, \(p=0.8\)) achieves substantial improvements in distant ranges, with increases of 11.42 and 12.59 percentage points in 60-80m and 80-100m, respectively. However, a drop of 11.58 points in the 40-60m range is linked to low diversity in the test set for this range, dominated by repetitive samples of a single stationary human instance. These repetitive samples, while well-segmented across frames, contribute to cumulative small errors, reducing the rIoU.

\begin{figure}[t]
\centering
\begin{subfigure}[b]{0.48\linewidth}
    \centering
    \includegraphics[width=\linewidth]{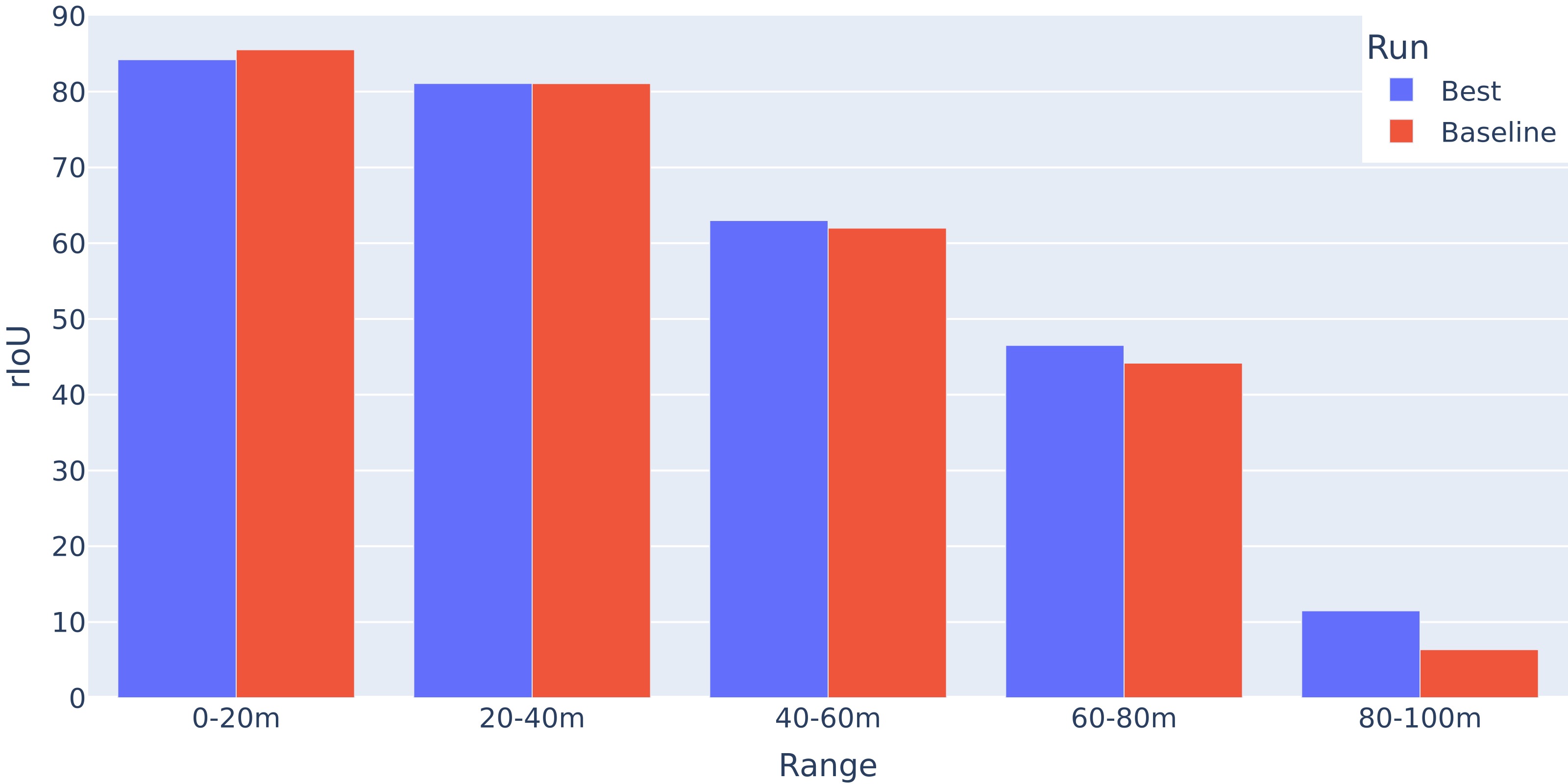}
    \caption{Class track: Range IoUs for baseline and TS model.}
    \label{fig:track_iou_test_set}
\end{subfigure}
\hfill
\begin{subfigure}[b]{0.48\linewidth}
    \centering
    \includegraphics[width=\linewidth]{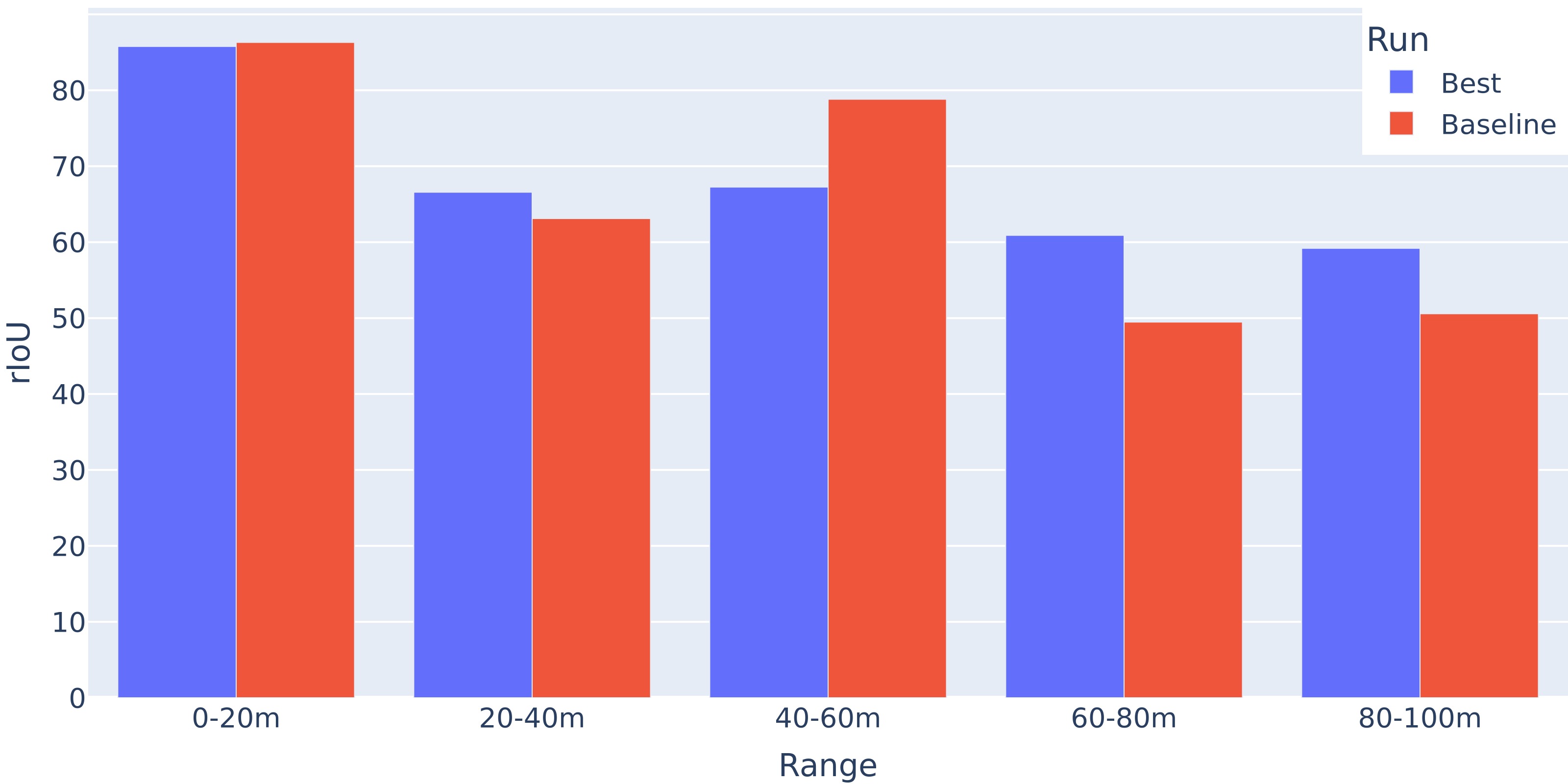}
    \caption{Class person: Range IoUs for baseline and PIP model.}
    \label{fig:person_iou_test_set}
\end{subfigure}
\caption{Comparison of range IoUs on the test set for baseline and the best-performing models. (a) Track sparsification (TS), (b) Person instance pasting (PIP online).}
\label{fig:range_iou_test_set}
\end{figure}

\subsubsection{Other Classes}
Table \ref{Table:IoU_comparison_test_set} summarizes the IoUs across all classes. The baseline model performs best overall for the person class, while the PIP online model achieves the highest track IoU. These results highlight that the methods are tailored to improve distant-range performance, leading to trade-offs in close-range inference. For the buffer stop class, all models show a near-complete IoU drop (from 93.86\% on validation to \textless 1\%  on the test set), due to overfitting to similar training-validation point clouds and poor generalization to the sparse test set.

\subsubsection{Discussion of Results}
The TS method enhances far-range performance while minimally impacting close-range inference, demonstrating its effectiveness in handling sparsified regions. Future work could explore variable DSDs for improved adaptability.

The PIP online method significantly boosts distant-range rIoUs but struggles in low-diversity regions such as 40-60m. Future improvements could include adapting the intensity field and creating a more diverse instance registry to enhance generalization.

\section{Conclusion}

The experiments on OSDaR23 validate the effectiveness of the proposed targeted data augmentations in improving segmentation performance at distant ranges, with minimal impact on close-range accuracy. The track sparsification and person instance pasting methods address key challenges in LiDAR-based semantic segmentation for autonomous trains.

Future work could integrate additional sensor data, such as RGB images, to leverage color information and enhance performance. Incorporating temporal data, as demonstrated in methods like MemorySeg \cite{liMemorySegOnlineLiDAR2023}, could further improve predictions by capturing motion and context. Additionally, exploring the inverse of track sparsification—densifying distant point clouds using techniques like \cite{youUpSamplingMethodLowResolution2023}—offers another avenue for enhancing segmentation in sparse regions.

These methods provide a solid foundation for advancing multimodal, temporal, and augmentation-driven approaches in semantic segmentation for autonomous train systems.

\section*{Acknowledgements}
This work has received funding from the German Federal Ministry for Economic Affairs and Climate Action (BMWK) under grant agreement 19I21039A.

{
    \small
    \bibliographystyle{ieeenat_fullname}
    \bibliography{references}
}

\end{document}